\documentclass[conference]{IEEEtran}
\IEEEoverridecommandlockouts
\usepackage{cite}
\usepackage{amsmath,amssymb,amsfonts}
\usepackage{algorithmic}
\usepackage{graphicx}
\usepackage{textcomp}
\usepackage{xcolor}
\def\BibTeX{{\rm B\kern-.05em{\sc i\kern-.025em b}\kern-.08em
    T\kern-.1667em\lower.7ex\hbox{E}\kern-.125emX}}

\begin{document}
\title{Multilingual Augmentation for Robust Visual Question Answering in Remote Sensing Images\\ 

}

\author{\IEEEauthorblockN{Zhenghang Yuan}
\IEEEauthorblockA{\textit{Data Science in Earth Observation} \\
\textit{Technical University of Munich (TUM)}\\
Munich, Germany \\
zhenghang.yuan@tum.de}
\and
\IEEEauthorblockN{Lichao Mou}
\IEEEauthorblockA{\textit{Data Science in Earth Observation}\\
\textit{Technical University of Munich (TUM)}\\
Munich, Germany\\
lichao.mou@tum.de}
\and
\IEEEauthorblockN{Xiao Xiang Zhu}
\IEEEauthorblockA{\textit{
Data Science in Earth Observation}\\
\textit{Technical University of Munich (TUM)}\\
Munich, Germany\\
xiaoxiang.zhu@tum.de}
}

\maketitle

\IEEEpubid{\begin{minipage}{\textwidth}\ \\[12pt] \centering
~ \\~ \\~\\  
\end{minipage}}


\begin{abstract}
Aiming at answering questions based on the content of remotely sensed images, visual question answering for remote sensing data (RSVQA) has attracted much attention nowadays. However, previous works in RSVQA have focused little on the robustness of RSVQA. As we aim to enhance the reliability of RSVQA models, how to learn robust representations against new words and different question templates with the same meaning is the key challenge. With the proposed augmented dataset, we are able to obtain more questions in addition to the original ones with the same meaning. To make better use of this information, in this study, we propose a contrastive learning strategy for training robust RSVQA models against diverse question templates and words. 
Experimental results demonstrate that the proposed augmented dataset is effective in improving the robustness of the RSVQA model. In addition, the contrastive learning strategy performs well on the low resolution (LR) dataset.

\end{abstract}


\section{Introduction}

With the development of remote sensing technology, there has been a significant increase in the number of available remote sensing images \cite{xiong2022earthnets}. However, these images are not as straightforward and easy to understand as natural images due to their special characteristic, such as overhead views and containing many types of land covers \cite{wang2018getnet,xiong2021benchmark}. Therefore, it is desirable to have something auxiliary to help understand remote sensing imagery. Natural language is a good choice since it can describe the content in the image. There are many tasks combining the image and natural language, such as image captioning, referring image segmentation, and visual question answering (VQA). Image captioning aims to provide descriptions about images \cite{sumbul2020sd}, referring image segmentation is to segment out objects by text from images \cite{ye2019cross}, and VQA is to answer the question in natural language based on the content of the image \cite{antol2015vqa}. Among them, VQA for remote sensing data (RSVQA) has attracted a lot of attention in recent years due to its usefulness for many tasks \cite{bazi2022bi, Chappuis2}.

\begin{figure}
	\centering
	\includegraphics[width=0.49\textwidth]{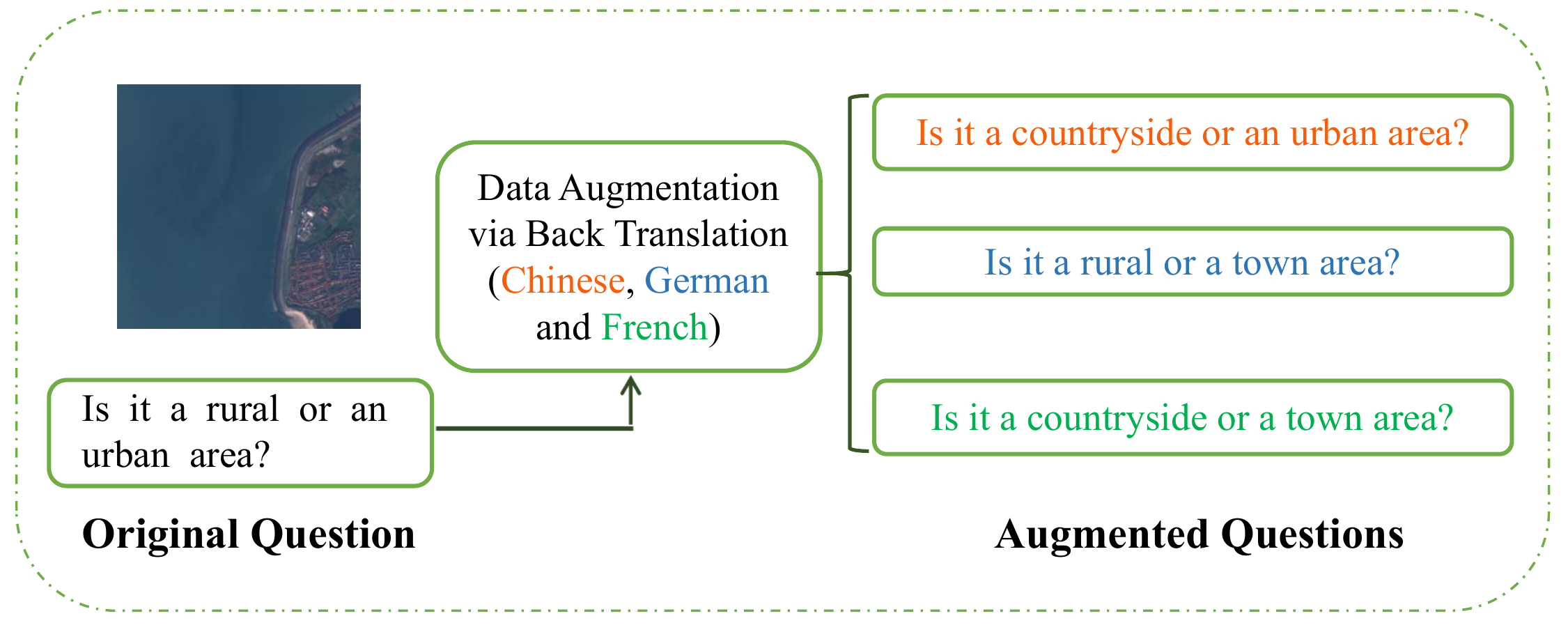}
	\caption{Examples of data augmentation via back translation.}
	\label{figure1}
\end{figure}

Given an image and a corresponding question, the purpose of RSVQA is to answer the question in natural language based on the image content. Specifically, the inputs are image-question pairs and the outputs are answers. This task was first introduced by the work \cite{lobry2020rsvqa}, where two datasets aiming at different applications were built. The authors used Convolutional Neural Networks (CNNs) for extracting visual features and a Recurrent Neural Network (RNN) for learning linguistic features to analyze datasets. Chappuis et al. \cite{Chappuis} proposed to convert the image to text and only used a language model to process visual context and question text. Besides, Lobry et al. \cite{Lobry2021} created a large-scale dataset called RSVQAxBEN containing nearly 15 million on top of the BigEarthNet data \cite{earthnet}, and this dataset provides much support for future RSVQA studies. Considering that questions have clearly different difficulty levels, Yuan et al. \cite{yuan2021from} proposed to train models with question samples in an easy-to-hard way. Recently, change detection-based VQA \cite{yuan2021CDVQA} is also introduced to help users understand land cover changes.


\begin{figure*}
	\centering
	\includegraphics[width=0.78\textwidth]{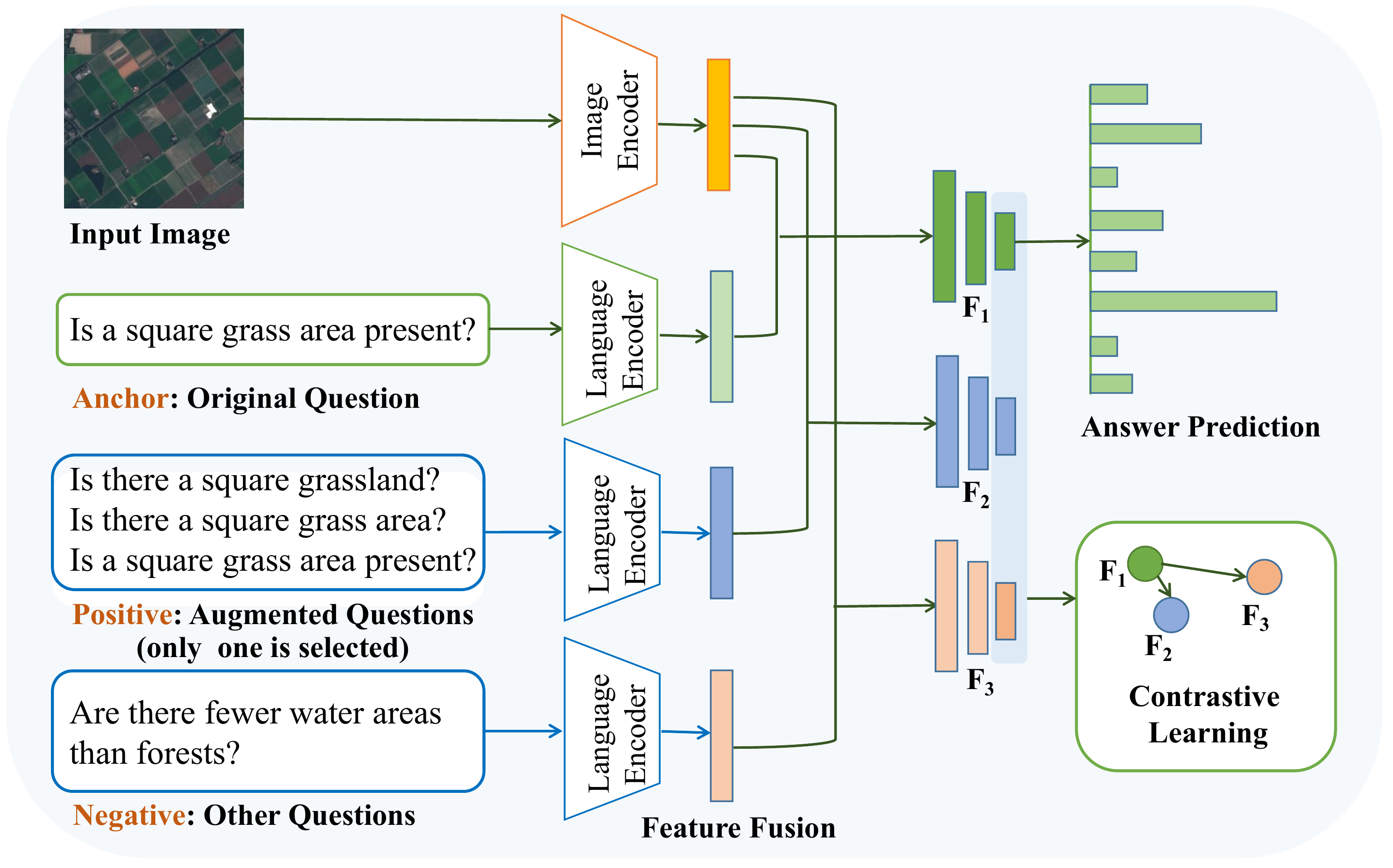}
	\caption{The overall architecture of the proposed method. }
	\label{figure3}
\end{figure*}

Almost all existing RSVQA datasets are automatically generated by the pre-designed templates. In the process of question generation, templates of questions are selected first, and then keywords are replaced to get different questions \cite{lobry2020rsvqa}. Notably, even though the number of questions is large, the diversity of the same type of questions is still limited by the number of templates. For instance, there are up to 14,758,150 image/question/answer triplets, but only 25 different questions for each image in the RSVQAxBEN dataset \cite{Lobry2021}. In this case, once the input questions contain new words and templates, the model may output wrong answers due to the limited extrapolation ability of the model. Therefore, improving the robustness of RSVQA is necessary to correctly answer new questions. In this study, we propose to exploit back translation to increase the number of question templates and words. Back translation is the process of translating translated language back to the source language. Specifically, as shown in Fig.\ref{figure1}, we use back translation as data augmentation via Chinese, German, and French to consider the language habits of different countries. By doing so, models will be more robust as it sees more augmented questions. In addition, we also develop contrastive learning \cite{contrastive} in this task to improve the generalization ability of models. The overall architecture of the proposed method is shown in Fig. \ref{figure3}.

\section{Data Augmentation}

Since natural language is very flexible, there are many different ways to express the same meaning. For example, when asking for the number of images, there can be expressions such as ``what is the number of images" and ``how many images are there", etc. Since it is impossible to cover all templates and synonyms in one dataset, models should be robust to unknown ones. Data augmentation is a way to increase the variety of templates and words. In addition, since end-users come from various countries, there are various syntaxes when different native speakers speak English with the same meaning. In this study, We use data augmentation from a multilingual perspective and propose to exploit back translation to increase the number of question templates and words. 
As shown in Fig. \ref{figure1}, back translation means the process of translating translated language back to the source language. For example, the language of the original questions in the RSVQA dataset is English. We first translated original questions into Chinese, German, and French, and then translate questions in these three languages back into English. In this case, back translation is the process of translating Chinese, German, and French questions back to new English questions. Notably, new English questions are the same meaning as the original English questions, but their sentences or words are likely to be different. By this means, we have different expressions of each question and thus augment the diversity of the original dataset.

In this study, we choose to augment the low-resolution dataset of RSVQA \cite{lobry2020rsvqa}, which is automatically generated using the OpenStreetMap data. To enhance the multilingual diversity of this dataset, a back translation technique is used in three different and widely-used languages: Chinese, German, and French. Due to different grammars and syntactic conventions, the diversity of questions in the original dataset can be significantly enhanced. To be more specific, we keep the original questions and increase their diversity by augmenting them with 66,909 questions back-translated from Chinese, 66,909 questions back-translated from German, and 66,908 questions back-translated from French. Finally, we construct a new dataset, termed augmented dataset with 277,958 questions, 77,232 answers, and 772 images.

\section{Methodology}
As we aim to enhance the reliability of RSVQA models, how to learn robust representations against new words and different question templates with the same meaning is the key challenge. With data augmentation, we are able to obtain more questions besides the original ones with the same meaning. To better utilize these augmented data, in this study, we propose a contrastive learning strategy for training robust RSVQA models against diverse question templates and words. The whole architecture of the proposed model is illustrated in Fig. \ref{figure3}. For each image in the batch, the original question, its augmented questions, and a question randomly selected from the batch are input to an RNN-based language encoder to obtain textual features. Meanwhile, the visual features are obtained with a CNN-based image encoder. Next, a feature fusion module is employed to fuse visual features with three textual features respectively. Formally, we assume that $F_1$ denotes the fusion result of the visual feature $F_v$ and the original language feature. $F_2$ is the fusion of $F_v$ and the language feature of a randomly selected augmented question, and $F_3$ represents the fusion output of $F_v$ and the language feature of a randomly selected question from the batch. Based on these definitions, we can formulate the triplet loss function for the contrastive learning strategy as follows:
\begin{equation}
\mathcal{L}(F_1, F_2, F_3)=\max \left\{d\left(F_{1_i}, F_{2_i}\right)-d\left(F_{1_i}, F_{3_i}\right)+m, 0\right\},
\end{equation}
where $d\left(x_i, y_i\right)=\left\|\mathbf{x}_i-\mathbf{y}_i\right\|_2$. $F_1$ is the original multi-modal feature of questions and images, $F_2$ is the new multi-modal feature of augmented questions and images, and $F_3$ is the reverse order of $F_1$ along the batch dimension. We use $m$ to denote the margin of the triplet loss. Basically, we set $F_1$ as the anchor sample, $F_2$ as the positive sample, and $F_3$ as the negative sample. By using the triplet loss, the model can learn similar representations for questions with diverse words and templates with the same meaning. 

Following previous works \cite{lobry2020rsvqa, Lobry2021}, we formulate RSVQA as a classification task, i.e., the answer is predicted from a pre-defined answer pool. Thus, the commonly-used cross-entropy loss is used for the model training. Specifically, for the original question $q_a$ and image $x$, the cross entropy loss can be defined by $\mathcal{L}(x,q_a,y)$. Similarly, the cross-entropy loss for the augmented question $q_p$ can be expressed as $\mathcal{L}(x,q_p,y)$. The final training loss $\mathcal{L}$ consists of the contrastive learning loss and two cross-entropy losses for the original and the augmented questions:
\begin{equation}
\mathcal{L}=\mathcal{L}(x,q_a,y)+\mathcal{L}(x,q_p,y)+\mathcal{L}(F_1, F_2, F_3).
\end{equation}

\section{Experiments and Discussion}
\subsection{Dataset}
We conduct experiments on the low resolution (LR) dataset \cite{lobry2020rsvqa} to prove the effectiveness of the proposed method. LR dataset is built based on the Sentinel-2
images with 10 m resolution. It consists of 772 images with the size of $256 \times 256$ and 77,232 question/answer pairs. 
To study the robustness of RSVQA models and validate the effectiveness of our method, we conduct experiments in three different settings: 1) training with the original set and testing on the original test set; 2) training with the original set and testing on the augmented set; 3) training with the augmented set and testing on the augmented set. In addition, we provide separate experimental results for the augmented dataset in three different languages. 

\subsection{Implementation Details} 
Pytorch is used to implement the proposed model. Specifically, we adopt the model from \cite{yuan2021from} as the baseline. For the training hyper-parameters, we use the Adam optimizer with a learning rate of 1e-5. The batch size is set to 280, and 150 epochs are used for the model training. As for contrastive learning loss, we set the margin parameter as 1 in this study.  

\begin{table}
    \centering
    \caption{Comparisons between our method and E2H \cite{yuan2021from} on the augmented dataset.}
    \resizebox{8.5cm}{!}{
        \begin{tabular}{c|c|c}
            \hline\hline
            Type & E2H \cite{yuan2021from} & Ours \\
            & Original $\rightarrow$ Augmented Set & Augmented $\rightarrow$ Augmented Set \\
            \hline
            Presence & 0.8471 & 0.9072 \\
            Count & 0.6167 & 0.7064 \\
            Comparison & 0.8193 & 0.8731 \\
            Rural/Urban & 0.8533 & 0.9100 \\
            AA & 0.7841 & 0.8474 \\
            OA & 0.7666 & 0.8323 \\
            \hline\hline
        \end{tabular}
    }
    \label{Table1}
\end{table}

\begin{table*}[]
\centering
\caption{Comparisons between our method and E2H on the augmented and original dataset. }
\resizebox{18cm}{!}{
\begin{tabular}{c|cccc}
\hline \hline
Type        & E2H\cite{yuan2021from},Original$\rightarrow$Chinese Set & E2H\cite{yuan2021from},Original$\rightarrow$German Set & E2H\cite{yuan2021from},Original$\rightarrow$French Set & E2H\cite{yuan2021from}, Original$\rightarrow$Original Set \\  \hline
Presence    & 0.8368                   & 0.848                   & 0.8565                  & 0.9011                     \\
Count       & 0.6199                   & 0.5826                  & 0.6477                  & 0.686                      \\
Comparison  & 0.7546                   & 0.8335                  & 0.8698                  & 0.8683                     \\
Rural/Urban & 0.8600                     & 0.8600                    & 0.8400                    & 0.9000                     \\
AA          & 0.7678                   & 0.7810                  & 0.8035                  & 0.8389                     \\
OA          & 0.7403                   & 0.7641                  & 0.7955                  & 0.8227    \\ \hline \hline                
\end{tabular} }
\label{Table2}
\end{table*}

\subsection{Results and Analyses}
We aim to learn robust multi-modal representations against new words and different question templates with the same meaning. To this end, a contrastive learning training strategy is proposed. In order to validate the effectiveness of our proposed method, we conduct experiments to compare the results on the original dataset and the multilingual-augmented dataset. For each sample in the augmented set, there are almost three more questions in addition to the original one with the same meaning. Note that ``original$\rightarrow$augmented set" denotes that the model is trained on the original dataset \cite{lobry2020rsvqa}, and tested on our augmented dataset, and the same applies to ``original$\rightarrow$original set" and ``augmented$\rightarrow$augmented set".    From the results in Table \ref{Table1}, we can see a clear performance drop if we test the model trained using the original dataset on the augmented one. The diversity in words and templates is the main reason for this degradation. For example, there is about a 7\% performance drop for the count question type and a roughly 5\% drop for the comparison questions. Fortunately, using our training method as well as the proposed augmented dataset can compensate for this performance degradation and even improve the model performance on the original set. Some visualization examples are in Fig. \ref{figure3}.

In order to study the effects of using different languages for back translation augmentation, we provide detailed performance for each individual language. The results are presented in Table \ref{Table2}. When the model is trained on the original dataset and tested on the Chinese back-translated dataset, we can see that the comparison-related questions are the most affected ones with an approximate 11\% performance drop. When it comes to German back-translated questions, there is a significant performance drop for the count-related questions. Interestingly, for the French back-translated questions, the performance degradation is relatively small. To sum up, a well-trained RSVQA model may exhibit decreased performance in real-world scenarios when users with different native languages are involved. This provides useful insights for future research toward robust RSVQA models.

\section{Limitation and Discussion}
Back translation can indeed generate diverse sentences for a language model's training, but it is important to note that it may not always guarantee the quality and correctness of the generated sentences. For example, a circular generation of texts can result in sentences that are correct but differ in meaning from the original text. 
To ensure the quality of generated sentences, manual evaluation is one possible solution, but it can be a time-consuming and expensive process. As a compromise solution, back translation is still a useful tool for generating diverse and creative variations of text although it may result in sentences with different meanings occasionally. 
Thanks to the emergence of advanced AI language models like ChatGPT, currently it is possible to generate high-quality and diverse text at scale, which can be a promising technique for enriching the original datasets and improving the performance of language models. In this context, ChatGPT can generate diverse sentences that are contextually relevant and grammatically correct, which can be useful to improve the generalizability of the trained RSVQA models.
\section{Conclusion}
The robustness and reliability of RSVQA models are important factors that need to be considered in real-world applications. To improve the robustness of existing RSVQA methods, we propose a multilingual-based data augmentation strategy to enhance RSVQA datasets. In the augmented dataset, for each original question, we obtain more questions with the same meaning using back-translation. To make better use of this information, on top of the augmented dataset, we also design a contrastive learning-based method for robust learning of multi-modal features. We show that the proposed augmented dataset can be used to improve the robustness of the RSVQA model. 

\begin{figure}
	\centering
	\includegraphics[width=0.45\textwidth]{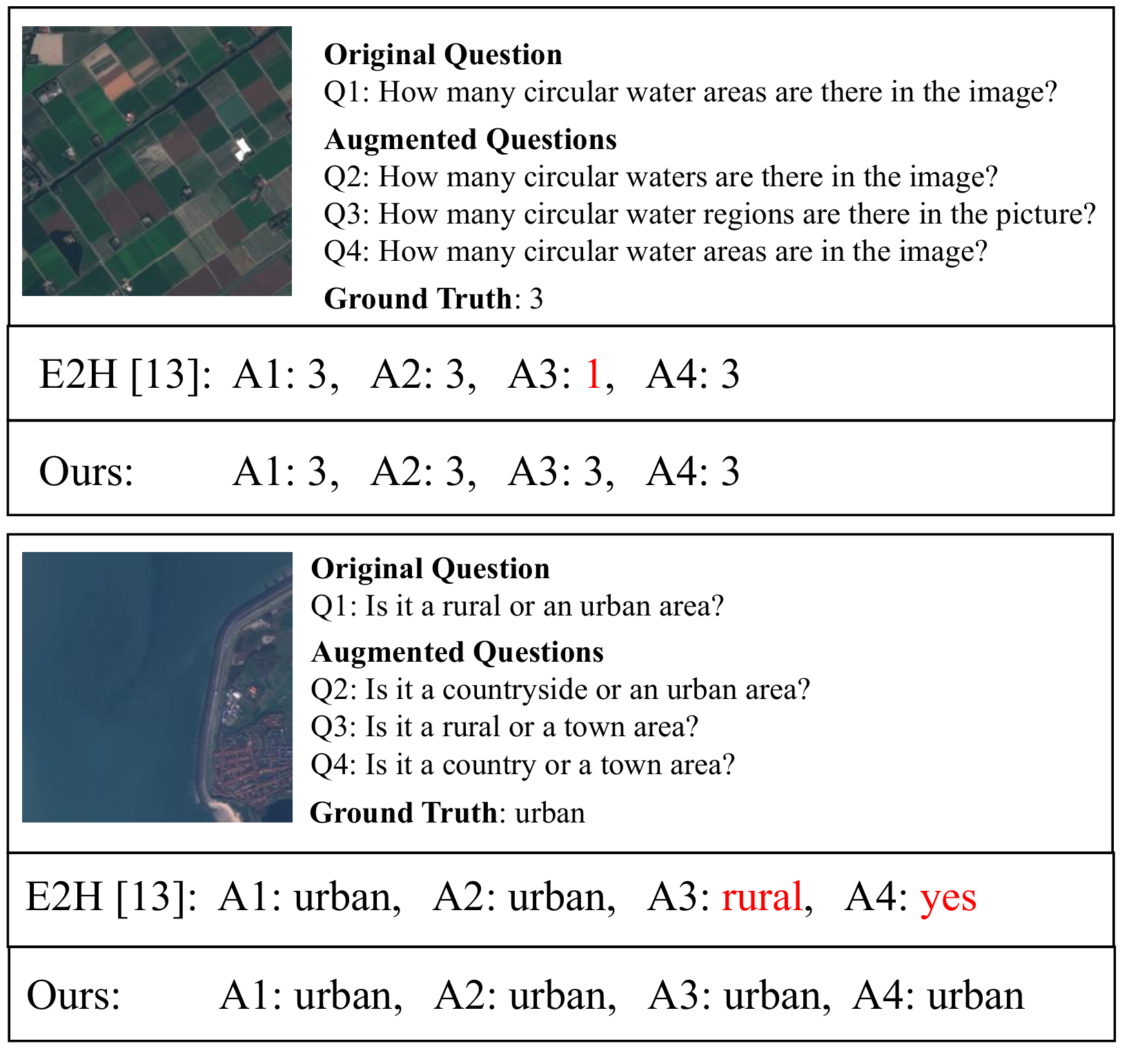}
	\caption{Visualization examples of experimental results for the model \cite{yuan2021from} and the proposed method. }
	\label{figure3}
\end{figure}

\vspace{12pt}
\end{document}